\documentclass{article}

\usepackage[preprint]{neurips_2024}

\usepackage{natbib} 
\usepackage[utf8]{inputenc} 
\usepackage[T1]{fontenc}    
\usepackage{hyperref}       
\usepackage{url}            
\usepackage{booktabs}       
\usepackage{amsfonts}       
\usepackage{nicefrac}       
\usepackage{microtype}      
\usepackage{xcolor}         
\usepackage{graphicx}
\usepackage{tabularx}
\usepackage{multirow}

\title{Position Paper On Diagnostic Uncertainty Estimation from Large Language Models: \\
Next-Word Probability Is Not Pre-test Probability}

\author{
Yanjun Gao$^{1,2*}$ \quad Skatje Myers$^2$ \quad Shan Chen$^3$ \quad Dmitriy Dligach$^4$ \\
\textbf{Timothy A Miller}$^5$ \quad
\textbf{Danielle Bitterman}$^6$ \quad \textbf{Guanhua Chen}$^2$ \\ \textbf{Anoop Mayampurath}$^2$ \quad \textbf{Matthew Churpek}$^2$ \quad \textbf{Majid Afshar}$^2$  \\
$^1$University of Colorado Anschutz \quad $^2$University of Wisconsin-Madison \\
$^3$Harvard Medical School \quad  
$^4$ Loyola University Chicago \\
\quad $^5$ Boston Childrens Hospital \quad $^6$ Dana Farber Cancer Institute \\
$*$\texttt{yanjun.gao@cuanschutz} \\
\texttt{\{symers, mchurpek, mafshar\}@medicine.wisc.edu}\\
\texttt{\{Shan.Chen, Timothy.Miller\}@childrens.harvard.edu} \\
\texttt{ddligach@luc.edu, DBITTERMAN@BWH.HARVARD.EDU} \\
\texttt{\{mayampurath, gchen25\}@wisc.edu}
}

\begin{document}

\maketitle

\begin{abstract}
  Large language models (LLMs) are being explored for diagnostic decision support, yet their ability to estimate pre-test probabilities, vital for clinical decision-making, remains limited. This study evaluates two LLMs, Mistral-7B and Llama3-70B, using structured electronic health record data on three diagnosis tasks. We examined three current methods of extracting LLM probability estimations and revealed their limitations. We aim to highlight the need for improved techniques in LLM confidence estimation. 
\end{abstract}

\section{Introduction}

Diagnosis in medicine is inherently complex and involves estimating the likelihood of various diseases based on a patient's presentation. This process requires integrating baseline information to establish pre-test probabilities during the initial hypothesis generation for a diagnosis, followed by iterative refinement as diagnostic test results become available~\citep{sox1989use,bowen2006educational} (Figure 1). Typically, clinicians rely on medical knowledge, pattern recognition and experience, enabling quick hypothesis generation of the initial diagnosis. However, this process is prone to cognitive biases, which can lead to diagnostic errors\citep{saposnik2016cognitive}. Analytic thinking, a more evidence-based process, is time-consuming and often impractical in fast-paced clinical environments. Although clinicians are taught to estimate a pre-test probability and apply test sensitivity and specificity, cognitive biases and heuristic-based thinking often lead to under- and overestimation of the pre-test probability and subsequent misdiagnoses~\citep{rodman2023artificial}.

The integration of Large Language Models (LLMs) in diagnostic decision support systems has garnered significant interest in addressing these challenges. Recent advancements, particularly with models like GPT-4, have demonstrated that LLMs can rival clinicians in generating differential diagnoses~\citep{kanjee2023accuracy,savage2024diagnostic}. However, LLMs often fail to explicitly convey uncertainty in the estimated probability of a diagnosis in their outputs. This is crucial in medicine; for example, an LLM might suggest an initial diagnosis of pneumonia, yet, a 20\% probability of pneumonia may have vastly different implications for a clinician compared to a 90\% probability. While GPT-4 has shown some potential for improvement over clinicians in predicting pre-test probability of certain conditions, overall performance is still suboptimal~\citep{rodman2023artificial,kanjee2023accuracy}. LLMs are not designed as classifiers that output probability distributions over specific outcomes; instead, they produce probability distributions over sequences of tokens. This raises the research question of how to map these token sequences to clinically meaningful probabilities, particularly for pre-test or post-test diagnosis probability estimation. Addressing this gap is crucial to avoid potential misinterpretations and mitigate the risk of automation bias in clinical settings. 

\begin{figure}
  \centering
  \includegraphics[scale=0.72]{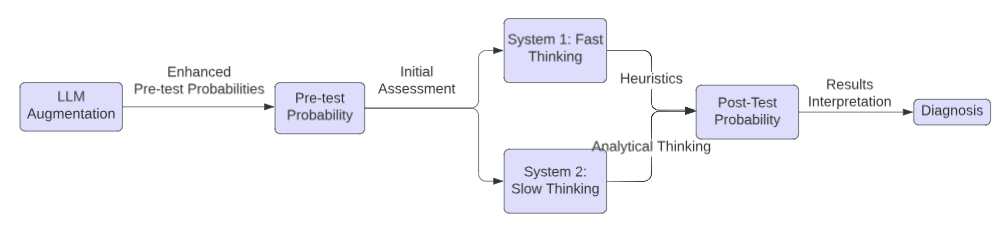}
  \vspace{-.12in}
  \caption{Process map in generating a diagnosis with the role of LLMs to augment human diagnostic reasoning.}
\end{figure}

The concept of uncertainty estimation for the generated text in LLMs is rooted in information theory with entropy, which measures the uncertainty of a probabilistic distribution to get next-word prediction. This process involves training the models to align their predictions with the actual distribution of the language they are trained on, resulting in the generation of convincing and coherent natural language. Existing literature investigates methods for extracting uncertainty estimation from LLMs, including token probabilities and verbalized probabilities (confidence)\citep{kapoor2024large,geng2024survey}. However, LLMs are known to suffer from the problem of unfaithful generation, where their outputs do not always accurately reflect their underlying knowledge or reasoning~\citep{hager2024evaluation,turpin2024language}. Further, while previous work~\citep{savage2024large} shows that sample consistency with embeddings could serve as uncertainty proxies, they evaluated on question-answering tasks, which is different from the real-world electronic health records setting. While LLMs may have general knowledge about disease prevalence from the pretraining corpora, such as Wikipedia, it remains uncertain whether they can translate general knowledge into patient-specific diagnostic reasoning and estimate pre-test probabilities, a question this paper aims to investigate. 

 We aimed to address this gap by evaluating the strengths and limitations of LLMs in pre-test diagnostic probability estimation. We conducted a detailed evaluation of two open-box LLMs: Mistral-7B-Instruct~\cite{jiang2023mistral} and Llama3-70B-chat-hf~\cite{touvron2023llama}, on the task of predicting pre-test diagnostic probabilities. These models were selected because they were available open source and adaptable through instruction-tuning. Unlike previous work exploring LLM medical uncertainty estimation on question-answering tasks or case reports~\citep{saposnik2016cognitive,hager2024evaluation,abdullahi2024learning}, our study was based on a set of annotated structured data in the electronic health records (EHRs) from a cohort of 660 patients at a large medical center in the United States. The task involves binary predictions for Sepsis, Arrhythmia, and Congestive Heart Failure (CHF), with positive class distributions of 43\%, 16\%, and 11\%, respectively. Ground truth diagnoses were annotated by expert physicians through chart reviews~\citep{Churpek2024}. The EHR data included vital signs, lab test results, nurse flow-sheet assessments, and patient demographics. We compared our results to an eXtreme Gradient Boosting (XGB) classifier that used the raw structured EHR data as input, representing the state-of-the-art in many clinical predictive applications~\citep{lolak2023comparing,govindan2024development}. EHR data included vital signs, structured nurse flowsheet assessments (i.e., mental status, mobility, etc.), and lab test results. We subsequently added patient demographics (sex, ethnicity, and race), encoded as categorical variables, to examine if such a setting could improve model performance. 

\section{Methods of Extracting Pre-test Probabilities from LLMs}

This section formulates the task as a binary diagnostic outcome classification using three methods. We benchmarked against XGB using raw features (baseline, Raw Data+XGB), correlating each method. We utilized a table-to-text method to convert structured EHR data into text. Specifically, we began this transformation by creating a template starting with “Hospitalized patient with age XX, systolic blood pressure YY …” where XX and YY represent the actual values from patient representation, as shown in Table~\ref{tab:mimic_template}. We then appended each clinical feature, its corresponding value, and its unit of measurement in the text. We refer readers to our recent paper, which provides a more detailed description of the table-to-text conversion process.~\citep{gao2024raw}.

\begin{table}
\scriptsize  
\begin{tabularx}{\columnwidth}{X}  
\toprule
Hospitalized patient of age \textit{[value]} getting worse has labs and vitals values of systolic blood pressure \textit{[value]} mmHg, diastolic blood pressure \textit{[value]} mmHg, oxygen saturation\textit{[value]} \%, body temperature \textit{[value]} celsius degree, ...  total protein \textit{[value]}, white blood cell \textit{[value]}. What are the diagnoses for this patient?  \\
\bottomrule
\end{tabularx}
\caption{\small The template for \textsc{narrative} serialization method for diagnosis prediction dataset.  }
\label{tab:mimic_template}
\end{table}

Our method of prompt development started with a prompt from previous literature that prompts GPT-4 for confidence estimation~\citep{rodman2023artificial}. We then made modifications according to the task descriptions and format requirements of the LLMs. To ensure that the LLMs would follow the format requirements, we tested the prompts with a few synthetic examples. 

\textbf{Token Logits} We prompted the LLM with a detailed description of the patient's condition and directly asked for a binary response: "Does the patient have {diagnosis}? A. Yes" or "B. No", indicating the presence or absence of sepsis. Specifically, the probability estimation was derived from the logits corresponding to these responses. We applied a softmax function yielding a normalized score for each option. We used a zero-shot setting for both LLMs. 

\textbf{Verbalized Confidence} This approach followed the previous study of GPT-4 on diagnostic probability estimation, and we used the same format of prompt~\citep{rodman2023artificial}. This set of prompts posed a more open-ended question to the LLM followed by a narrative description of patient representation: "How likely is it for the patient to have {diagnosis}?" The LLM will provide a percentage score, which we utilized as the probability of positive diagnosis. This approach allowed us to assess the model's ability to generate and verbalize probability assessments in a natural language format without further binarization of the results. We used zero-setting for both LLMs. Instead of applying a cut-off to categorize the predictions, we evaluated the raw probability estimates directly using AUROC scores, Pearson Correlation and Expected Calibration Errors (ECE). 

\textbf{Feature-based Calibrators} For the feature-based calibrators, we applied max pooling to the last layer of the LLMs, forming the embedding representation in 4096 and 8192 feature spaces for Mistral and Llama3-70B, respectively. These embeddings were then fed into an XGB classifier. All XGB classifiers were trained using a five-fold cross-validation on the 660 patient data.

\textbf{Training details }
All experiments involving XGB classifiers were trained under a 5-fold cross-validation setting. On each fold, we employed grid search for parameters using another five-fold to select the best hyperparameters. The parameter grid included $n\_estimators$ set to [50, 100, 250, 500], $max\_depth$ ranging from [2, 5, 10, 15, 20], $learning$ $rate$ values of [0.005, 0.01, 0.05, 0.1], and $min\_child\_weight$ values of [1, 2, 3].

\section{Results} 

\begin{figure}
    \centering
    \includegraphics[width=0.8\linewidth]{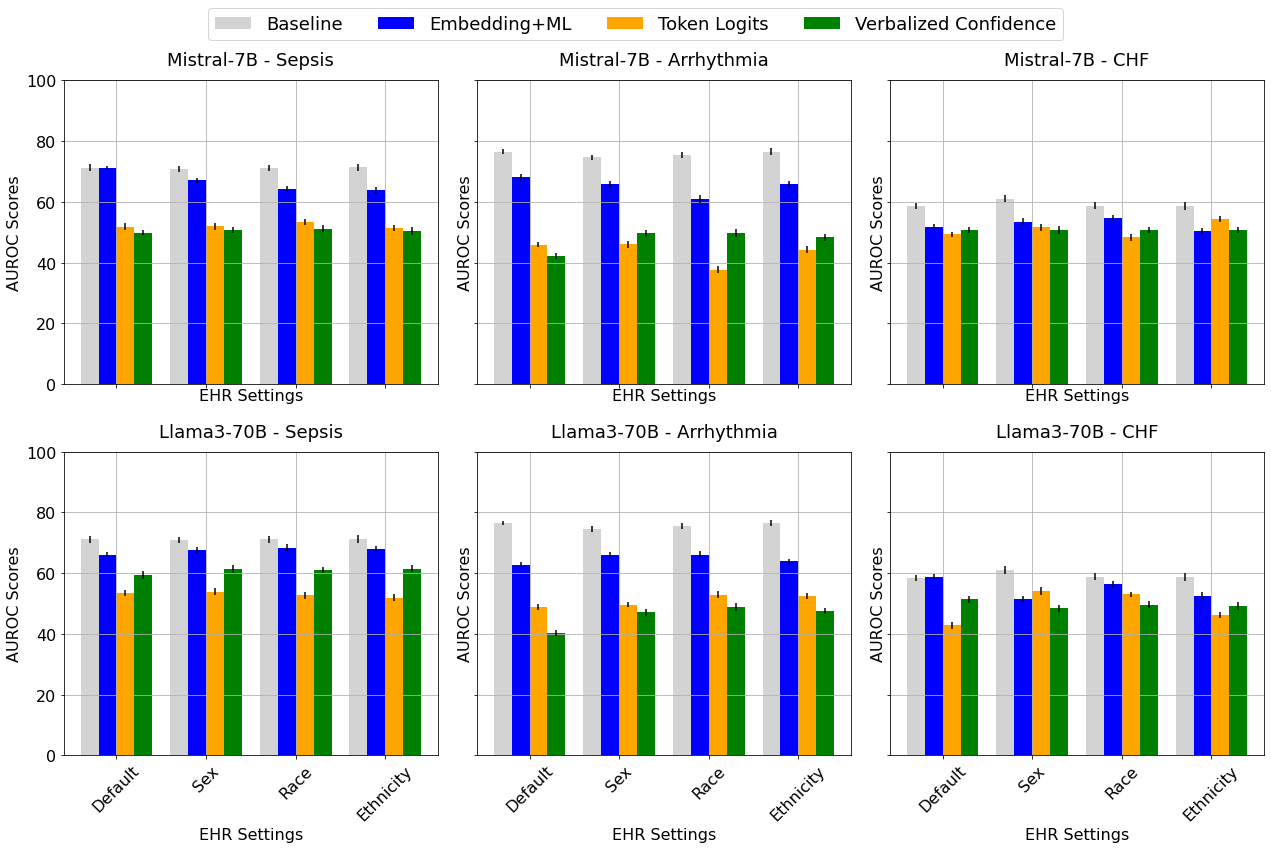}
    \vspace{-.1in}
    \caption{\small Area under the receiver operating characteristic curve (AUROC) scores from both LLMs using different EHR demographics input settings, across the diagnoses prediction of Sepsis, Arrhythmia, and Congestive Heart Failure (CHF).}
    \label{fig:auroc}
\end{figure}

Figure~\ref{fig:auroc} illustrates the results of the AUROC from LLMs to predict Sepsis, Arrhythmia, and CHF. The LLM Embedding+XGB method consistently outperformed the other LLM-based methods. Particularly for Sepsis, it achieved nearly the same AUROC score as the baseline Raw Data+XGB. The Token Logits (mean AUROC: 49.9 with 95\% CI: 47.8-51.9) and Verbalized Confidence (mean AUROC: 50.9 with 95\% CI: 48.7-53.1) methods exhibited marginal performance, generally not surpassing the baseline XGB classifier. The inclusion of demographic variables (sex, race, ethnicity) changed the AUROC scores, for instance, by as much as 7.22 for Mistral embedding on Sepsis prediction (71.1 on default setting vs 63.9 on ethnicity). However, the direction and consistency of these changes varied depending on the specific context and data included. 

\begin{figure}
    \centering
    \includegraphics[width=0.8\linewidth]{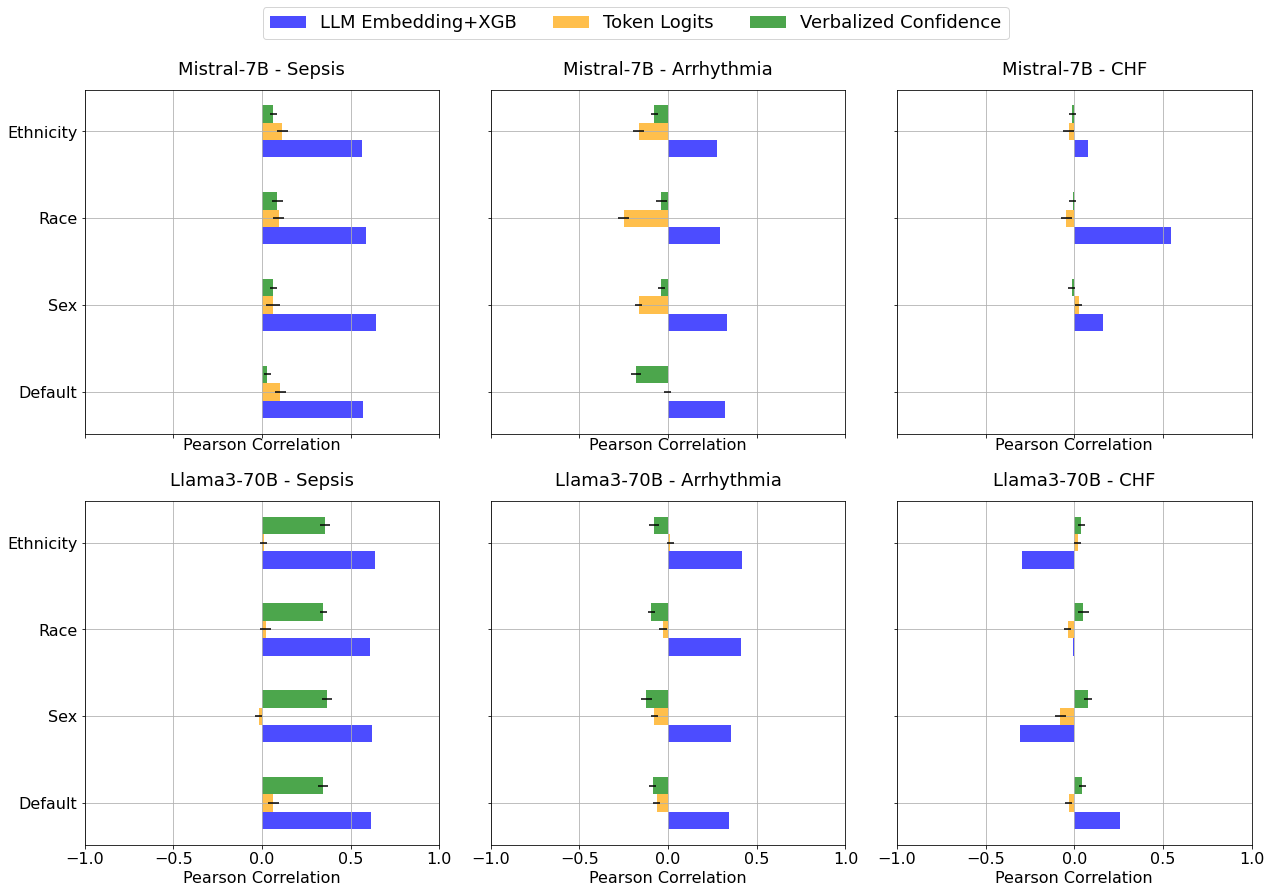}
    \vspace{-.1in}
    \caption{\small Pearson correlations between the LLM predicted pre-test probabilities and the baseline model (Raw Data+XGB) predicted probabilities.  }
    \label{fig:pearson}
\end{figure}

\begin{figure}
    \centering
    \includegraphics[width=0.8\linewidth]{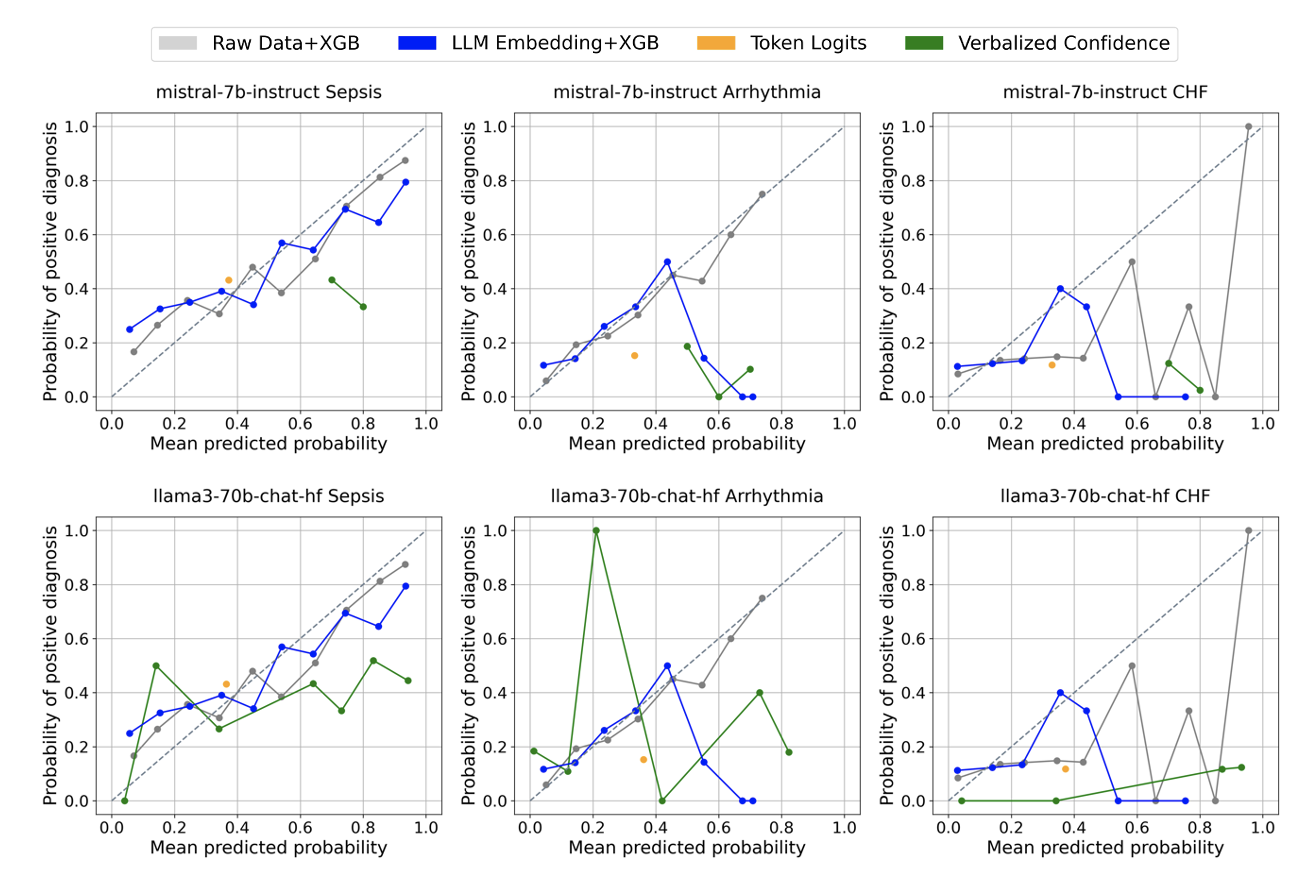}
    \vspace{-.1in}
    \caption{\small Calibration curves on the four probability estimation methods, using Mistral-7B-Instruct and Llama3-70B-Chat-hf on Default patient demographic setting. }
    \label{fig:calibration}
\end{figure}

\begin{table}[ht!]
\small 
\centering
\begin{tabular}{c c c c c}
\toprule
\textbf{Method} & \textbf{None } & \textbf{Sex } & \textbf{Race} & \textbf{Ethnicity} \\ 
& Sepsis, Arrhy., CHF & Sepsis, Arrhy., CHF & Sepsis, Arrhy., CHF & Sepsis, Arrhy., CHF \\ 
\midrule
\multicolumn{5}{c}{Baseline} \\ \midrule
\textbf{Raw+XGB} & 0.09, 0.02, 0.10 & 0.07, 0.02, 0.10 & 0.09, 0.02, 0.10 & 0.10, 0.02, 0.11 \\ \midrule
\multicolumn{5}{c}{Mistral-7B-Instruct} \\  
\midrule
\textbf{Token Logits} & 0.07, 0.18, 0.21 & 0.13, 0.21, 0.25 & 0.13, 0.21, 0.25 & 0.13, 0.21, 0.25 \\ 
\textbf{Verb. Conf.} & 0.27, 0.43, 0.59 & 0.31, 0.35, 0.58 & 0.27, 0.35, 0.59 & 0.27, 0.36, 0.58 \\ 
\textbf{Emb+XGB} & 0.09, 0.06, 0.13 & 0.14, 0.09, 0.09 & 0.11, 0.11, 0.11 & 0.06, 0.06, 0.11 \\ 
\midrule
\multicolumn{5}{c}{Llama3-70B-Chat} \\ \midrule
\textbf{Token Logits} & 0.07, 0.21, 0.25 & 0.06, 0.22, 0.25 & 0.07, 0.21, 0.25 & 0.06, 0.22, 0.25 \\ 
\textbf{Verb. Conf.} & 0.28, 0.25, 0.77 & 0.24, 0.04, 0.01 & 0.24, 0.04, 0.04 & 0.24, 0.04, 0.03 \\ 
\textbf{Emb+XGB} & 0.11, 0.05, 0.07 & 0.06, 0.09, 0.17 & 0.06, 0.04, 0.09 & 0.06, 0.06, 0.11 \\ 
\bottomrule
\end{tabular}

\caption{\small ECE Results across the four methods for pre-test probabilities estimation methods, over three diagnosis prediction tasks with patient demographic settings.}
\label{tab:ece_results}
\end{table}

Figure~\ref{fig:pearson} reports the Pearson correlation coefficients between the predicted probabilities from LLM-based uncertainty estimation methods and those from the XGB classifier for three diagnoses across different demographics. When correlating the LLMs’ positive class probabilities with the baseline results, the token logits and verbalized confidence methods had more variable correlations, often no correlation or negative correlation, suggesting less alignment with the baseline XGB predicted probabilities. On the contrary, the LLM embedding+XGB method consistently showed strong positive correlations across all tasks and settings with the baseline XGB classifier, indicating its predictions were closely aligned with the baseline XGB classifier.

Figure~\ref{fig:calibration} presents the calibration curve of all models on the default setting (no demographic variable). Poor calibration was observed, especially from Token Logits and Verbalized Confidence. Probabilities predicted by the Token Logits always fell between ranges of 0.323-0.337. Table~\ref{tab:ece_results} further highlights the ECE scores of each method, with notable differences observed across the various biases and diagnoses. For instance, the Verbalized Confidence (Verb. Conf) method tends to exhibit higher ECE values, indicating poorer calibration, especially in the CHF prediction task, while Raw+XGB generally shows more consistent performance across different demographic settings. These results reflect the lack of robustness of these methods for uncertainty estimation, highlighting a significant gap in uncertainty estimation for medical decision-making. Although these methods are used in literature to assess uncertainty in predicting the next word~\citep{xiongcan}, predicting pre-test probabilities requires an understanding of risk based on real-world patient data and disease prevalence, knowledge that LLMs often lack. The introduction of demographic variables complicates the predictive power of these models further due to inherent biases present in LLMs, which may not be trained on a diverse set of patient characteristics, making them susceptible to social biases.

\section{Discussion and Conclusion}
The LLM Embedding+XGB method demonstrated competitive performance compared to the state-of-the-art XGB baseline classifier under specific conditions, such as Sepsis, and exhibited the strongest correlation among the methods tested. However, this result is not surprising given that both methods rely on training a classifier. In contrast, purely LLM-based methods, such as Token Logits (next-word probability) and Verbalized Confidence, were found to be unreliable for risk estimation. Their performance, evaluated through AUROC scores, Pearson Correlation, and calibration curves, deteriorated significantly when diagnosing conditions with lower prevalence, raising concerns about the accuracy of pre-test probabilities derived from these models. The results were consistent across both the Mistral-7B and Llama3-70B models. Additionally, results varied with different demographic characteristics, reinforcing existing concerns about bias in LLMs~\citep{zack2024assessing}. While the LLM Embedding+XGB method showed promise in generating pre-test probabilities, overall, LLM-based probability estimation methods did not achieve the same level of performance as raw tabular data in an XGB model. This underscores the necessity for further optimization of LLM methods to produce uncertainty estimations that align more closely with established and reliable methods.

Overall, our findings demonstrate the inability of LLMs to provide reliable pre-test probability estimations for specific diseases and highlight the need for improved strategies to incorporate numeracy into diagnostic decision support systems and reduce the impact of bias on LLM performance. This remains a major gap to fill before we can enter a new era of diagnostic systems that integrate LLMs to augment healthcare providers in their diagnostic reasoning. 
To address these limitations, future work should explore hybrid approaches that integrate LLMs with numerical reasoning modules or calibrated embeddings, enhancing their capacity for accurate uncertainty estimation, particularly for low-prevalence conditions. Additionally, bias mitigation strategies, such as layer-wise probing and targeted regularization, could help ensure fairer predictions across demographic groups. These advancements would bring us closer to safe and effective diagnostic systems that integrate LLMs to support clinicians in clinical decision-making. 

\section*{Competing Interests}
There is no competing interests to be declared. 
 
\section*{Funding Acknowledgments}

This work is supported by the National Library of Medicine (NLM) under award R00LM014308 02 (PI: Gao), NLM award R01LM012973 05 (PI: TM, DD, MA), National Heart, Lung, and Blood Institute (NHLBI) under award R01HL157262 04 (PI: MC), NIH-USA awards U54CA274516 01A1 and R01CA294033 01 (PI: DB; SC), NSF DMS-2054346 and the University of Wisconsin School of Medicine and Public Health through the Wisconsin Partnership Program (Research Design Support: Protocol Development, Informatics, and Biostatistics Module, PI: GC).

\bibliographystyle{apalike}
\bibliography{reference}

\end{document}